\documentclass{article}
\usepackage{ijcai19}
\usepackage{amsmath,amssymb,amsfonts}
\usepackage{times}
\usepackage{soul}
\usepackage{url}
\usepackage[hidelinks]{hyperref}
\usepackage[utf8]{inputenc}
\usepackage[small]{caption}
\usepackage{graphicx}
\usepackage{booktabs}
\usepackage{bm}
\usepackage{enumerate}
\usepackage{textcomp}
\usepackage{calc}
\usepackage{subfigure}
\usepackage{wrapfig}
\usepackage[table]{xcolor}
\usepackage{colortbl}
\usepackage{framed}
\usepackage[framemethod=TikZ]{mdframed}
\usepackage{multirow}
\usepackage{mathrsfs}
\usepackage{diagbox}
\usepackage[ruled]{algorithm2e}
\newlength\colomnWidth
\newtheorem{proof}{\textbf{Proof}}
\newtheorem{property}{\textbf{Property}}

\title{
    TraceWalk: Semantic-based Process Graph Embedding for Consistency Checking
}

\author{
    Chen Qian$^{1}$\and Lijie Wen$^{1}$\footnote{Corresponding Author}\And Akhil Kumar$^{2}$\\
    \affiliations
    $^1$School of Software, Tsinghua University, Beijing, China\\
    $^2$Smeal College of Business, Penn State University, University Park, USA\\
    \emails
    qc16@mails.tsinghua.edu.cn, wenlj@tsinghua.edu.cn, akhilkumar@psu.edu
}

\begin{document}

\maketitle

\begin{abstract}
    Process consistency checking (PCC), an interdiscipline of natural language processing (NLP) and business process management (BPM), aims to quantify the degree of (in)consistencies between graphical and textual descriptions of a process. However, previous studies heavily depend on a great deal of complex expert-defined knowledge such as alignment rules and assessment metrics, thus suffer from the problems of low accuracy and poor adaptability when applied in open-domain scenarios. To address the above issues, this paper makes the first attempt that uses deep learning to perform PCC. Specifically, we proposed {\tt TraceWalk}, using semantic information of process graphs to learn latent node representations, and integrates it into a convolutional neural network (CNN) based model called {\tt TraceNet} to predict consistencies. The theoretical proof formally provides the PCC's lower limit and experimental results demonstrate that our approach performs more accurately than state-of-the-art baselines.
\end{abstract}

\section{Introduction}
Process knowledge, also called “how-to-do-it” knowledge, is the knowledge related to the execution of a series of interrelated tasks \cite{extracting_control_flow}. Nowadays, many organizations maintain huge process knowledge in various representations, including graphical and textual descriptions. The graphical descriptions of process knowledge (\textbf{process graphs}) have been found to be better suited to express complex execution logic of a process in a more comprehensive manner. By contrast, some stakeholders, especially workers who actually execute the process, have difficulties reading and interpreting process graphs and thus prefer textual descriptions (\textbf{process texts}) \cite{supporting_process}.

\begin{figure}[t]
    \centering
    \includegraphics[width=0.95\colomnWidth]{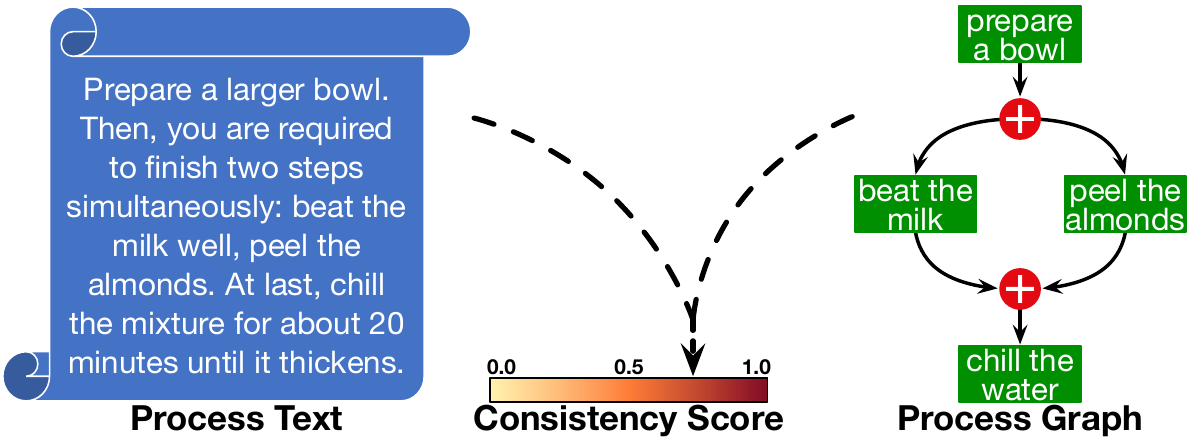}
    \caption{Illustration of the consistency checking task.}
    \label{fig:scenario}
    \end{figure}

Despite these benefits, the usage of multiple descriptions of the same process would lead to considerable inconsistencies inevitably when these formats are maintained or changed by independent organizations \cite{detecting_inconsistencies_between,comparing_textual_descriptions}. PCC aims to effectively measure the degree of (in)consistencies, correlate and retrieve the procedural data with different forms \cite{detecting_inconsistencies_between}. To illustrate, in Figure \ref{fig:scenario}, given a process graph\footnote{Crawled from {\tt \url{https://cookingtutorials.com}}, a famous cooking tutorial site.} and a process text, the consistency value between them is expected to be quantified. PCC not only makes machines understand procedural knowledge more intelligently, but also helps with the implementation of process correlation and retrieval. However, this task is challenging due to the ambiguity and variability of graphical and linguistic expressions.

Some prior studies have been conducted to automatically measure consistencies of graph-text pairs \cite{detecting_inconsistencies_between,comparing_textual_descriptions,aligning_textual_and_graphical,aligning_textual_and_model}. Traditional methods are heavily depend on effective alignment strategies, i.e., focusing mainly on how to align graph nodes and text sentences. The commonly used alignment strategies include best-first searching \cite{detecting_inconsistencies_between,comparing_textual_descriptions} and integer linear programming \cite{aligning_textual_and_model}. Therefore, such methods face two main issues. First, using traditional tagging or parsing depends heavily on the performance of existing NLP tools or systems. Second, requiring many domain-specific alignment rules and assessment metrics causes the weak generalization and adaptability. Thus, they are unable to be applied to open-domain or open-topic scenarios.

Unlike previous works, we design a CNN based model called {\tt TraceNet} to perform information understanding and consistency assessment without defining complex alignment rules and assessment metrics. Like \cite{learning_text_similarity,siamese_recurrent_architectures}, we consider the PCC task as a text similarity evaluation task. We propose {\tt TraceWalk} for learning semantic-based latent representations of process graph nodes. {\tt TraceWalk} uses local information obtained from truncated process graph traces to learn latent representations by treating traces as the equivalent of sentences. To fully perceive local textual information and reduce the number of model parameters, we design a siamese architecture with word-level convolution filter which is an architecture for non-linear metric learning with similarity information. It naturally learns representations that embody the invariance and selectivity through explicit information about similarity between pairs of objects. Specifically, we train a feedforward network which uses graphical and textual features as input, fuses them by hidden layers, and outputs consistency values. We proved the mathematical expectation value of random predicting which provides the PCC's lower limit, and compared our method with several state-of-the-art baselines. Experimental results demonstrate that {\tt TraceNet} consistently outperforms the existing methods.

In summary, this paper makes the following contributions:

\begin{enumerate}[$\bullet$]
    \item To the best of our knowledge, this work is the first attempt that brings deep learning in consistency checking. Supported by automatic feature extraction, our method can better understand process graphs and texts without using complex NLP tools and defining alignment rules.
    \item We propose a semantic-based process graph embedding technique called {\tt TraceWalk}, from which we can obtain semantic-based node vectors of process graphs effectively.
    \item We prove the lower limit of the PCC task and conduct extensive experiments. The extensive experiments yield consistently superior results. In addition, we make them publicly available as benchmark datasets for relevant fields.
\end{enumerate}

\section{Related Work}
Recently, some NLP techniques are applied to address a variety of use cases in the context of BPM. This includes a variety of works that focus on process graph labels, for example by annotating process graph elements and correcting linguistic guideline violations \cite{natural_language_in}, investigating the problem of mixing graphical and textual languages \cite{when_language_meets}, or resolving lexical ambiguities in process graphs labels \cite{automatic_detection_and,dealing_with_behavioral}. Other use cases involve process texts generation \cite{structural_descriptions} or process graph extraction \cite{extracting_control_flow}. However, these methods have been found to produce inaccurate results and require extensive manual participation.

Along this line, alignment-based PCC methods designed various alignment rules between process graphs and texts. We summarize and show their used procedures in Table \ref{tab:frameworks}. \cite{detecting_inconsistencies_between} (language-analysis based) set out to create an action-sentence correspondence relation between an action of a process graph and a sentence of a process text through linguistic analysis, similarity computation and best-first searching. \cite{comparing_textual_descriptions} (language-analysis based) extended previous work in order to detect missing actions and conflicting orders. Thus can detect inconsistencies in a much more fine-granular manner. \cite{aligning_textual_and_graphical} (manual-feature based) extended the linguistic analysis and encoded the problem of computing an alignment as the resolution of an integer linear programming problem. \cite{aligning_textual_and_model} (manual-feature based) extracted features that correspond to important process-related information and used so-called predictors to detect if a provided model-text pair is likely to contain inconsistencies. These methods require language analysis, manual feature extraction and sentence similarity computation etc. They heavily depend on existing language analysis tools, and always cause the weak generalization and adaptability.

Other related work includes graph embedding \cite{deepwalk,node2vec,a_comprehensive_survey} and text similarity \cite{convolutional_neural_network,learning_to_rank,siamese_recurrent_architectures,learning_text_similarity,an_enhanced_convolutional}. Graph embedding converts graph data into a low dimensional space in which the graph structural information and graph properties are maximumly preserved. Text similarity learns a similarity function between pairs of sentences or documents. These methods usually need a large parallel corpus for learning an end-to-end model.

Different from these studies, we aim to measure the consistency value between graph-text pairs without defining any alignment rule and assessment metric.


\begin{table}[t]
    \setlength{\abovecaptionskip}{0pt}
    \setlength{\belowcaptionskip}{0pt}
    \centering
    \small
    \caption{The comparison of existing and our proposed methods. LAB: language-analysis based methods. MFB: manual-feature based methods. OPM: our proposed method. $\star$: requiring expert-defined knowledge. $\bullet$: including. $\circ$: excluding. \\}
    \label{tab:frameworks}
    \begin{tabular}{|c|c|c|c|}
        \hline
        \textbf{Traditional Steps} & \textbf{LAB} & \textbf{MFB} & \textbf{OPM} \\
        \hline
        Language Analysis & $\bullet$ & $\bullet$ & $\circ$ \\
        \hline
        $\star$ Manual Feature Extraction & $\bullet$ & $\bullet$ & $\circ$ \\
        \hline
        $\star$ Sentence Similarity Computation & $\bullet$ & $\bullet$ & $\circ$ \\
        \hline
        $\star$ Find Optimal Correspondence & $\bullet$ & $\bullet$ & $\circ$ \\
        \hline
        $\star$ Inconsistency Assessment Metric & $\bullet$ & $\bullet$ & $\circ$ \\
        \hline
        \end{tabular}
    \end{table}

\section{Methodology}

\begin{figure*}[t]
    \centering
    \includegraphics[width=1.99\colomnWidth]{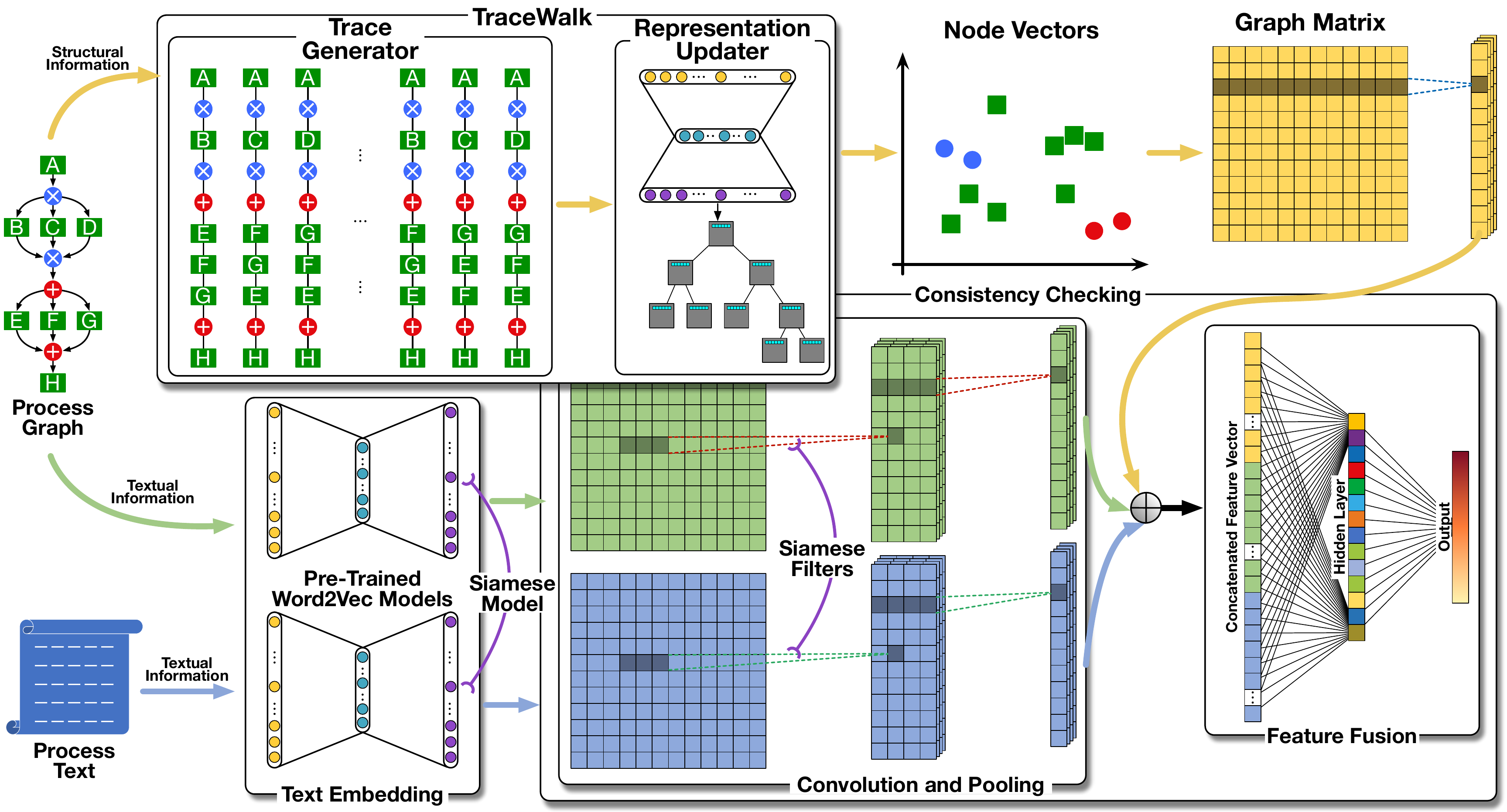}
    \caption{An overview of {\tt TraceNet}. Given a process graph and a process text, {\tt TraceWalk} aims to obtain semantic node vectors, text embedding to obtain textual word vectors, and consistency checking to learns and predicts the (in)consistencies of the input graph-text pair.}
    \label{fig:framework}
    \end{figure*}

\subsection{Task Definition and Overview}
We first define the studied problem as follows. Given a labeled process graph $\mathcal{G}=(V,E,\hbar)$ and a process text $\mathcal{T} = \langle S^1, S^2, \cdots, S^n \rangle$ where $V$ is a set of nodes with textual labels, $E \subseteq V \times V$ is a set of edges, $\hbar$ is a mapping function that maps each $v \in V$ to a specific type: \emph{activity} or \emph{gateway}, $S^i$ ($1 \le i \le n$) is a natural language sentence and $n$ is the number of sentences in $\mathcal{T}$. The goal is to learn a consistency function $\Phi$ which measures a normalized real number $\mathcal{Y} \in [0.0,1.0]$ denoting the degree of (in)consistencies between $\mathcal{G}$ and $\mathcal{T}$, i.e., $\Phi(\mathcal{G},\mathcal{T}) = \mathcal{Y}$. In particularly, the consistency value $1.0$ ($0.0$) denotes that $\mathcal{G}$ and $\mathcal{T}$ express a totally identical (different) process.

The overview of our designed {\tt TraceNet} is presented in Figure \ref{fig:framework}, including three parts:
\begin{enumerate}[$\bullet$]
    \item \textbf{Semantic Embedding} ({\tt TraceWalk}): transforming the execution semantics of $\mathcal{G}$ to a semantic vector space.
    \item \textbf{Text Embedding}: transforming the text semantics of $\mathcal{G}$ and $\mathcal{T}$ to text vector spaces.
    \item \textbf{Consistency Checking}: learning the non-linear consistency mapping function and predicting new examples.
\end{enumerate}


\subsection{{\tt TraceWalk}}
Since traditional graph embedding techniques \cite{deepwalk} ignore node types and execution semantics, we propose a process graph embedding technique called {\tt TraceWalk} which can effectively learns latent representations of nodes with executing semantics. {\tt TraceWalk} learns representation for vertices from a stream of semantic paths and uses optimization techniques originally designed for language modeling. It satisfies the following characteristics:
\begin{enumerate}[$\bullet$]
    \item \textbf{Adaptability} - new ordering relations should not require repeating learning when process graphs are evolving.
    \item \textbf{Low dimensional} - When labeled data is scarce, low-dimensional models generalize better, and speed up convergence and inference.
    \item \textbf{Continuous} - We require latent representations in continuous space. In addition to providing a nuanced view of execution semantics, a continuous representation allows more robust classification.
\end{enumerate}

Specifically, {\tt TraceWalk} consists of two main components: a process graph trace generator and a representation updater, aiming to generate semantic paths of process graphs and learn latent node vectors from them respectively.

\subsubsection{Process Graph Trace Generator}
In this section, we briefly introduce some basic patterns in BPM, leading to a new concept: process graph trace.

\begin{figure}[htbp]
    \centering
    \subfigure[SQ]{
        \label{sfig:pattern_sequence}
        \includegraphics[height=0.18\colomnWidth]{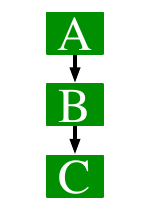}
    }
    \hspace{0.0cm}
    \subfigure[XOR]{
        \label{sfig:pattern_xor}
        \includegraphics[height=0.18\colomnWidth]{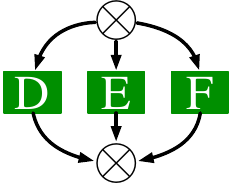}
    }
    \hspace{0.0cm}
    \subfigure[AND]{
        \label{sfig:pattern_and}
        \includegraphics[height=0.18\colomnWidth]{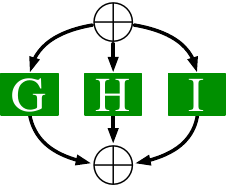}
    }
    \hspace{0.0cm}
    \subfigure[OR]{
        \label{sfig:pattern_or}
        \includegraphics[height=0.18\colomnWidth]{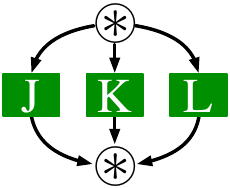}
    }
    \caption{Four basic structures of process graphs. SQ pattern belongs to sequential pattern, while XOR, AND and OR patterns not.}
    \label{fig:patterns}
  \end{figure}

As Figure \ref{fig:scenario} shows, a process graph contains two types of nodes: \emph{activities} (denoted by rectangles) and \emph{gateways} (denoted by circles). Activities represent points in a process where tasks are performed, and gateways control how activities run \cite{_bpmn}. Process graphs impose certain restrictions on the relationships between \emph{gateway} elements. In particular, in a process graph, each split gateway must have a corresponding join gateway. There are four basic types of patterns in process graphs: SQ pattern, XOR pattern, AND pattern, and OR pattern, as shown in Fgure \ref{fig:patterns}.

A process graph can be composed inductively based on these four types or patterns. When a process $\mathcal{G}=(V,E,\hbar)$ is executed, it has to abide by four basic \textbf{execution semantics}:

\begin{enumerate}[$\bullet$]
    \item SQ pattern: if an edge $\langle A,B \rangle \in E$, $B$ can be executed only \textbf{after} executing $A$.
    \item XOR pattern: only \textbf{one} of the XOR bounded activities can be executed.
    \item AND pattern: \textbf{all} of the AND bounded activities should be executed.
    \item OR pattern: \textbf{some} of the OR bounded activities can be executed.
\end{enumerate}

A process graph trace (PGT) is a complete node path following execution rules. For example, $\langle A,B,C \rangle$ is a PGT of Figure \ref{sfig:pattern_sequence}, $\langle \otimes,D,\otimes \rangle$ is a PGT of Figure \ref{sfig:pattern_xor}, $\langle \oplus,G,H,I,\oplus \rangle$ is a PGT of Figure \ref{sfig:pattern_and}, and $\langle \circledast,J,K,\circledast \rangle$ is a PGT of Figure \ref{sfig:pattern_or} ($\langle \circledast,K,J,\circledast \rangle \neq \langle \circledast,J,K,\circledast \rangle$). Note that, as for AND and OR patterns, a stack of activities can be executed simultaneously, and their execution orders depend on practical conditions. If process graph consists of a sequence of basic patterns, then, the PGT set of $\mathcal{G}$ is the cartesian product of each pattern's PGT set.

PGT generator sets out to generate a complete PGT set for each process graph. Note that no matter a process graph contains loop structure or not, we randomly generate an enough PGT set with an pre-defined upper limit. In addition to capturing semantical information, using random traces as the basis for our algorithm gives us a desirable property: relying on information obtained from random traces makes it possible to accommodate small changes in the graph structure without the need for global recomputation. We can iteratively update the learned model with new random traces from the changed region in time sub-linear to the entire graph.

\subsubsection{Representation updater}
As for a PGT $\tau^i=\langle \tau^i_1,\tau^i_2,\cdots,\tau^i_{|\tau^i|} \rangle$, it can be thought of as a sentence, i.e., a sequence of words. Therefore, we present a generalization of language modeling to explore the process graph through a stream of PGTs. We use a one-hidden-layer network to learn each node's representation.

More formally, given a sequence of traces $\langle \tau^1,\tau^2,\cdots,\tau^N \rangle$ where $\tau^i$ is a sequence of node $\langle \tau^i_1,\tau^i_2,\cdots,\tau^i_{|\tau^i|} \rangle$, the objective of the {\tt TraceWalk} model is to maximize the average log probability:

\begin{equation}
    \frac{1}{N} \sum_{i=1}^{N} \Big( \frac{1}{|\tau^i|} \sum_{t=1}^{|\tau^i|} \sum_{-c \le j \le c, j \ne 0} \log p(\tau_{t+j}^i|\tau_t^i) \Big)
\end{equation}

where $c$ is the size of the training context (larger $c$ results in more training examples and thus can lead to a higher accuracy, at the expense of the training time). Besides, we use softmax function to define $p(\tau_{t+j}^i|\tau_t^i)$:

\begin{equation} \label{equ:softmax}
    p(\tau_O^i|\tau_I^i) = \frac{\exp{({\bm{v}_{\tau_O^i}'}^\top {\bm{v}_{\tau_I^i}}})} {\sum_{v=1}^V \exp{({\bm{v}_{\tau_v^i}'}^\top {\bm{v}_{\tau_I^i}}})}
\end{equation}

where $\bm{v}_w'$ and $\bm{v}_w$ are the “input” and “output” vector representations of $w$, and $V$ is the number of graph nodes.

In Equation \ref{equ:softmax}, computing the probability distribution is expensive, so instead, we will factorize the conditional probability using hierarchical softmax \cite{hierarchical}. We assign the nodes to the leaves of a Huffman Binary Tree (HBT), turning the prediction problem into maximizing the probability of a specific path in HBT (see the upper part in Figure \ref{fig:framework}). For each, we map each node $v_j$ to its current representation vector $\Phi(v_j) \in \mathbb{R}^d$. If the path to node $w_k$ is identified by a sequence of tree nodes $\langle n_0,n_1,\cdots,n_{\lceil \log |V| \rceil} \rangle$, ($n_0=\text{root}, n_{\lceil \log |V| \rceil}=w_k$) then:

\begin{equation}
    p(w_k|\Phi(w_I)) = \prod_{l=1}^{\lceil \log |V| \rceil} \frac{1}{1+\exp(-[\![n_l]\!] {\bm{v}_{n_l}'}^\top\bm{v}_{w_I})}
\end{equation}

where $[\![n_l]\!]$ be $1$ if $n_l$ is the left child node of $n_{l-1}$ and $-1$ otherwise. This reduces the computational complexity of calculating probability distribution from $O(|V|)$ to $O(\log|V|)$.

\subsection{Text Embedding}
As for the textual information in process graphs and process texts, we pre-train other {\tt Word2vec} model on Google’s public Text8 corpus\footnote{{\tt \url{ https://code.google.com/archive/p/word2vec/}}} \cite{distributed_representations}. Finally, we concatenate all the word vectors to form the text features. If the short text is not long enough to up to $M$, we will pad $0$ in the end.

\subsection{Consistency Checking}
\subsubsection{Convolution and Pooling}
We employ multiscale word-level filters to capture local information of different length in a sentence \cite{multiperspective}. Let $\bm{x_1^n}$ refer to the concatenation of vectors $\bm{x_1}, \bm{x_2}, \cdots, \bm{x_n}$. The convolutional layer involves a set of filters $\bm{w} \in \mathbb{R}^{h \times k}$, which is solely applied to a window of $h$ to produce a new feature map $\bm{v}=
  \begin{bmatrix}
    \begin{bmatrix}
      \sigma (\bm{w_i} \bm{x}_1^h + b_i)\\
      \sigma (\bm{w_i} \bm{x}_2^{h+1} + b_i)\\
      \cdots \\
      \sigma (\bm{w_i} \bm{x}_{n-h+1}^n + b_i)\\
    \end{bmatrix}
  \end{bmatrix}
$, where $\sigma(\cdot)$ is a non-linear function and $b_i$ is a bias term.

To extract the most important features (max value) within each feature map and to make an accurate prediction, we employ max-pooling mechanism $\hat{v}=max(\bm{v})$.

\subsubsection{Feature Fusion}
After concatenating the pooled outputs $\mathcal{S}, \mathcal{L} , \mathcal{T}$: $\bm{V} = \mathcal{S} \oplus \mathcal{L} \oplus \mathcal{T}$, it is input into three fully connected feature fusion layer: $\bm{o_i} = softmax(\bm{W_2} \cdot (\bm{W_1} \cdot \bm{V} + \bm{b_1}) + \bm{b_2})$, where $\bm{W}$ and $\bm{b}$ are parameters of a network. $softmax$ operation aims to obtain the probability distribution on each type $t\in[1,T]$: $p_k=\frac{\exp (o_t)}{\sum_{i=1}^T \exp (o_i)}$, where $T$ is the number of a classification task. Finally, the output neuron stands for the corresponding predicted value.

\subsubsection{Siamese Mechanism}
Siamese networks \cite{siamese_recurrent_architectures,learning_text_similarity} are dual-branch networks with tied weights, i.e., they consist of the same network copied and merged with an energy function.

There are two {\tt Word2Vec} models $M$ and $N$ and two set of filters $F^a=<f_1^a,f_2^a,\cdots,f_m^a>$ and $F^b=<f_1^b,f_2^b,\cdots,f_n^b>$ which are used to perform convolution in two textual inputs, but we solely focus on siamese architectures with tied weights such that:

\begin{equation}
    \begin{cases}
        M \equiv N\\
        m \equiv n\\
        f_i^a \equiv f_i^b, \forall i \in [1,2,\cdots,m]
    \end{cases}
\end{equation}

\subsection{Loss Function}
We use quantile loss function to train our end-to-end model, when given a set of training data $x_i$; $y_i$; $e_i$, where $x_i$ is the $i$-th training example to be predicted, $y_i$ is the ground-truth value and $e_i$ is the predicted output. The goal of training is to minimize the loss function:

\begin{equation}
    J(\theta,\gamma) = \frac{1}{M} \Big( \sum_{i:e_i \le y_i} \gamma|e_i-y_i| + \sum_{i:e_i > y_i} (1-\gamma)|e_i-y_i| \Big)
\end{equation}

where $M$ is the number of training samples; $\gamma \in [0.0,1.0]$ is the linear punishment factor between positive and negative examples, and it regresses to mean absolute error if $\gamma=0.5$.

\section{Experiments}
\subsection{Datasets}
\begin{figure}[htbp]
    \centering
    \includegraphics[width=0.95\colomnWidth]{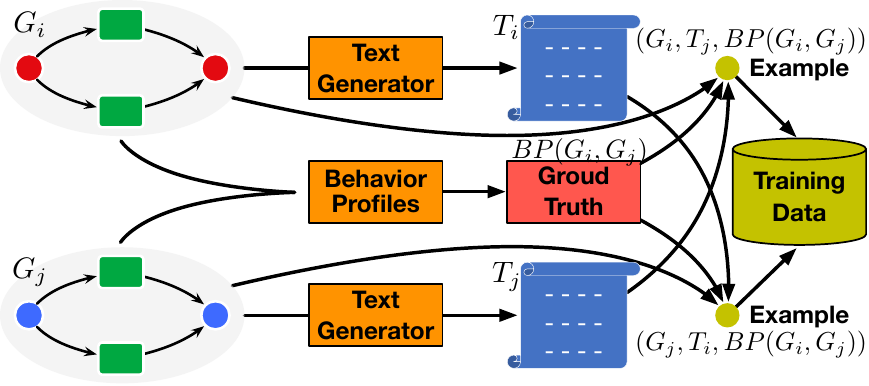}
    \caption{The main procedure of automatically generating training examples from process graphs.}
    \label{fig:dataset}
    \end{figure}

Although there are a great number of process graphs and texts, manually labeling consistency values between them is a time-consuming and error-prone task. Therefore, we refer to the work of \cite{aligning_textual_and_graphical,aligning_textual_and_model} and focus on automatically generating labels. As Figure \ref{fig:dataset} shows, when given two process graphs $G_i$ and $G_j$, we employ the state-of-the-art process translator {\tt Goun} \cite{structural_descriptions} to generate the corresponding process texts $T_i$ and $T_j$, and we use Behavior Profile (BP), an algorithm to evaluate similarities between two process graphs \cite{efficient_consistency_measurement}, to obtain gold ground truths $BP(G_i,G_j)$. By doing these, we can get two training examples $(G_i,T_j,BP(G_i,G_j)$ and $(G_j,T_i,BP(G_i,G_j)$. In this way, we can obtain enough graph-text datasets when given graph-graph datasets without manually time-consuming and error-prone labeling. The statistics of our graph datasets can be seen in Table \ref{tab:statistics} (train ratio is set to $80\%$).

\textbf{Randomly Generated Graphs (RGG)}. We use BeehiveZ\cite{beehivez} and random algorithm to generate random graphs with diverse structures and varying node numbers.

\textbf{Structured Process Repository (SPR)}. We adopted structured process graphs collected from the world’s third largest independent software manufacturer: SAP\footnote{{\tt \url{https://www.sap.com}}}.

\textbf{Industrial Petri Nets (IPN)}. We use Petri nets of three industrial enterprises: DG\footnote{{\tt \url{https://www.dbc.com.cn}}}, TC\footnote{{\tt \url{http://www.crrcgc.cc}}} and IBM\footnote{{\tt \url{https://www.ibm.com}}} \cite{structural_descriptions}.

\textbf{Academic BPMN Models (ABM)}. We use BPMN models of three academic sources: \cite{structural_descriptions}, \cite{aligning_textual_and_graphical} and BAI\footnote{{\tt \url{https://bpmai.org}}}.

\begin{table}[htbp]
  \setlength{\abovecaptionskip}{0pt}
  \setlength{\belowcaptionskip}{0pt}
  \centering
  \caption{Statistics of the datasets. \#: The average number of; $SMR$: The ratio of structured models to all models.}
  \label{tab:statistics}
  \begin{tabular}{|c|c|c|c|c|}
    \hline
     & \textbf{RGG} & \textbf{SPR} & \textbf{IPN} & \textbf{ABM} \\
     \hline
     Type & Generated & Industry & Industry & Academic \\
     \hline
     \# Graph & 2284 & 394 & 1222 & 602 \\
     \hline
     \# Node & 23.0 & 7.7 & 50.4 & 37.5 \\
     \hline
     \# $SMR$ & 47\% & 100\% & 76\% & 42\% \\
    \hline
    \end{tabular}
\end{table}

\begin{table*}[t]
    \setlength{\abovecaptionskip}{0pt}
    \setlength{\belowcaptionskip}{0pt}
    \centering
    \caption{Experimental results of horizontal comparisons (FCCM, MACO, ILP, PIIMC) and ablation analysis ({\tt TraceNet-$\varnothing$}, {\tt TraceNet-D}) on three tasks. {\tt TraceNet-$\varnothing$}, {\tt TraceNet-D} respectively denote variants of {\tt TraceNet} by removing {\tt TraceWalk} and replacing {\tt TraceWalk} with {\tt DeepWalk}. The best results are highlighted in bold.}
    \label{tab:mae}
    \begin{tabular}{|c||cccc||cccc||cccc|}
        \hline
        \multirow{2}*{Method} & \multicolumn{4}{c||}{\textbf{Task1: PCC on Activities}} & \multicolumn{4}{c||}{\textbf{Task2: PCC on Gateways}} & \multicolumn{4}{c|}{\textbf{Task3: PCC on All}} \\
        & \textbf{RGG} & \textbf{SPR} & \textbf{IPN} & \textbf{ABM} & \textbf{RGG} & \textbf{SPR} & \textbf{IPN} & \textbf{ABM} & \textbf{RGG} & \textbf{SPR} & \textbf{IPN} & \textbf{ABM} \\
        \hline
        \hline
        FCCM & 0.102 & 0.202 & 0.115 & 0.188 & 0.470 & 0.290 & 0.503 & 0.539 & 0.347 & 0.303 & 0.269 & 0.389 \\
        MACO & 0.076 & 0.163 & 0.069 & 0.126 & 0.432 & 0.241 & 0.469 & 0.476 & 0.295 & 0.284 & 0.215 & 0.333 \\
        ILP & 0.062 & 0.126 &  \textbf{0.043} & 0.068 & 0.398 & 0.195 & 0.444 & 0.432 & 0.250 & 0.269 & 0.164 & 0.295 \\
        PIIMC & \textbf{0.053} & 0.117 &  \textbf{0.043} & 0.059 & 0.390 & 0.189 & 0.434 & 0.423 & 0.241 & 0.266 & 0.156 & 0.290 \\
        \hline
        \hline
        {\tt TraceNet-$\varnothing$} & 0.083 & 0.173 & 0.106 & 0.128 & 0.526 & 0.364 & 0.140 & 0.338 & 0.248 & 0.198 & 0.184 & 0.304 \\
        \hline
        {\tt TraceNet-D} & 0.067 & 0.142 & 0.072 & 0.068 & 0.137 & 0.157 & 0.059 & 0.072 & 0.137 & 0.165 & 0.082 & 0.099 \\
        \hline
        {\tt TraceNet} & 0.058 & \textbf{0.112} & 0.058 & \textbf{0.056} & \textbf{0.118} & \textbf{0.131} & \textbf{0.055} & \textbf{0.063} & \textbf{0.109} & \textbf{0.134} & \textbf{0.064} & \textbf{0.067} \\
        \hline
    \end{tabular}
    \end{table*}

\subsection{Implementation Details}
Our method was implemented with {\tt Tensorflow} framework. The semantic embedding size and the word embedding size are set to 100. The maxlength of the siamese filters is set to 100, and the number of them is 128. Our model contains has 128 hidden cells for each feature fusion layer. Moreover, we used a sigmoid function as an activation unit and the Adam optimizer \cite{adam} with a mini-batch size of 128. The models were run at most 10K epochs. The learning rate is 0.0002 and we decrease it to 0.0001 after 7K epochs. All the matrix and vector parameters are initialized with uniform distribution in $[-\sqrt{6/(r+c)},\sqrt{6/(r+c)}]$, where $r$ and $c$ are the numbers of rows and columns in the matrices \cite{understanding_the_difficulty}. We empirically set the hyper-parameter $\gamma=0.7$ in loss function. The training stage averagely took half an hour on a computer with three GeForce-GTX-1080-Ti GPUs.

\subsection{Baselines}
We compare our method with four state-of-the-art baselines:
\begin{enumerate}[$\bullet$]
  \item A language-analysis based method (FCCM) \cite{detecting_inconsistencies_between}, which presents the first method to automatically identify inconsistencies between a process graph and a corresponding process text.
  \item An extended language-analysis based method (MACO) \cite{comparing_textual_descriptions}, which considers missing activities and conflicting orders.
  \item A manual-feature based method (ILP) \cite{aligning_textual_and_graphical}, which focuses on alignment by encoding the search as a mathematical optimization problem.
  \item A manual-feature based method (PIIMC) \cite{aligning_textual_and_model}, which is grounded on projecting knowledge extracted from graphs and texts into an uniform representation that is amenable for comparison.
\end{enumerate}

\subsection{The Mathematical Expectation of Randomly Predicting}

In our study, we further prove the randomly predicted value ($\mathscr{L}$) of the PCC task which can be the lower limit of efficient PCC methods i.e., an effective PCC method is bounded from below by $\mathscr{L}$.

\begin{property}
    Given a pross graph $\mathcal{G}$ and a process text $\mathcal{T}$, their gold consistency value is $p$. The randomly predicted result $\mathscr{L}$ by a random consistency checker is $q$. Then, we have the equation: $\mathscr{L}=E(|p-q|)=\frac{1}{3}$.
\end{property}

\begin{figure}[htbp]
    \centering
    \includegraphics[width=0.55\colomnWidth]{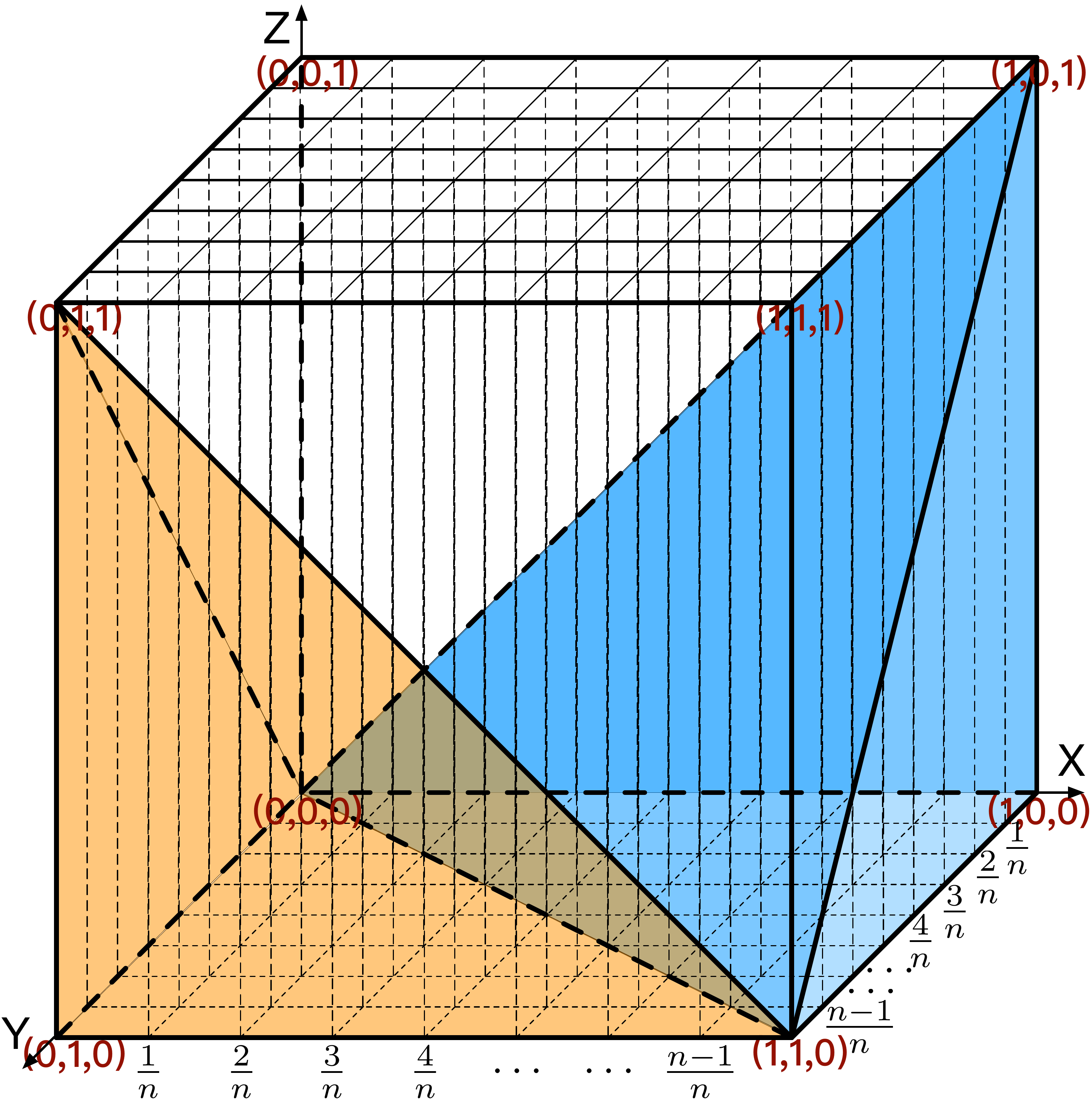}
    \caption{Decomposed cube for proving the random PCC value.}
    \label{fig:proof}
\end{figure}

\begin{proof}
    Suppose that we have a PCC cube ($1 \times 1 \times 1$) shown in Figure \ref{fig:proof}. Let $X,Y,Z$ be the gold value $p$, the randomly predicted value $q$ and $|p-q|$ respectively. Then, we decompose the cube into finite $n^2$ ($\frac{1}{n} \times \frac{1}{n} \times 1$) bricks. For a brick $e$ located on $(\frac{i}{n},\frac{j}{n})$, its height (predicted value) is $\frac{|i-j|}{n}$. Hence:
\end{proof}

\begin{equation}
    \begin{aligned}
        E(|p-q|) = & \lim_{n \rightarrow \infty} \Big( \frac{1}{n} \times \frac{1}{n} \sum_{i=1}^n \sum_{j=1}^n |\frac{i}{n}-\frac{j}{n}| \Big) \\
        = & \lim_{n \rightarrow \infty} \Big( \frac{2}{n^3} \sum_{i=1}^n \sum_{j=1}^i (i-j) \Big) \\
        = & \lim_{n \rightarrow \infty} \Big(\frac{n(n+1)(2n+1)}{6n^3} - \frac{n(n+1)}{2n^3} \Big) \\
        = & \lim_{n \rightarrow \infty} \Big(\frac{1}{3} - \frac{1}{2n^2} \Big) \\
        = & \frac{1}{3}
    \end{aligned}
\end{equation}

\subsection{Comparison with Baselines}
In order to test and verify whether or not our method can deal with complex graph-text pairs from single type of nodes, we compare it with baselines on three tasks: embedding only activities (Task1), gateways (Task2) and all nodes (Task3). The comparison results are shown in Table \ref{tab:mae}.

\textbf{Horizontal comparison}. We observe that although ILP and PIIMC produces slightly better accuracies on Task1, the difference is not significant (0.053 vs. 0.058, 0.043 vs. 0.058). The main reason is that FCCM, MACO, ILP and PIIMC employ many aligning features and strategies instead of learning, which tends to over-fit some specific datasets. Besides, we also can conclude that FCCM, MACO, ILP and PIIMC do not work well on Task2 and Task3, even producing values worse than randomly predicting (0.539 and 0.476 etc.), while {\tt TraceNet} achieves the minimum errors among all methods on IPN and ABM. In summary, {\tt TraceNet} decreases the error by up to 0.352, 0.159, 0.448, and 0.476 respectively, which demonstrates its efficacy. 

\textbf{Vertical comparison}. We would like to evaluate the impact so we test {\tt TraceNet} by removing {\tt TraceWalk} (denoted by “{\tt TraceNet-$\varnothing$}”) or replacing it with {\tt DeepWalk} (denoted by “{\tt TraceNet-D}”). The ablation results show that {\tt TraceWalk} (semantic embedding) is an effective mechanism in process embedding compared with {\tt DeepWalk} (structure embedding).

In summary, we can conclude that our method achieves significant improvement over the other baselines for almost all benchmarks and tasks, which demonstrates the effectiveness of our proposed {\tt TraceWalk} mechanism and the attempt of applying deep learning in PCC.






\section{Conclusion and Future Work}
In this paper, we explore deep learning to evaluate the consistency value of a graph-text pair. We empirically demonstrated that our model outperforms state-of-the-art baselines and also exhibited the proof of randomly predicted value theoretically. In the future, it would be interesting to explore the feasibility of learning more knowledge from graphs and texts.

\bibliographystyle{named}
\bibliography{Consistency_IJCAI}

\end{document}